\documentclass[conference]{IEEEtran}
\IEEEoverridecommandlockouts
\usepackage{cite}
\usepackage{comment}
\usepackage{titlesec}
\usepackage{gensymb}

\usepackage{multirow}
\usepackage{amsmath,amssymb,amsfonts}
\usepackage{algorithmic}
\usepackage{graphicx}
\usepackage{gensymb}
\usepackage{textcomp}
\usepackage{xcolor}
\usepackage{url}
\usepackage{subcaption}
\usepackage{booktabs}
\usepackage{tabularx}
\def\BibTeX{{\rm B\kern-.05em{\sc i\kern-.025em b}\kern-.08em
    T\kern-.1667em\lower.7ex\hbox{E}\kern-.125emX}}
\begin{document}

\title{Contextual Autonomy Evaluation of Unmanned Aerial Vehicles in Subterranean Environments
}

\author{\IEEEauthorblockN{Ryan Donald \hspace{1cm} Peter Gavriel \hspace{1cm} Adam Norton \hspace{1cm} S. Reza Ahmadzadeh}
\IEEEauthorblockA{\textit{PeARL lab and NERVE Center} \\
\textit{University of Massachusetts Lowell}\\
Lowell, USA \\
Ryan\_Donald@student.uml.edu \; Peter\_Gavriel@uml.edu \; Adam\_Norton@uml.edu \; Reza\_Ahmadzadeh@uml.edu}

}
\maketitle

\begin{abstract}
In this paper we focus on the evaluation of contextual autonomy for robots. More specifically, we propose a fuzzy framework for calculating the autonomy score for a small Unmanned Aerial Systems (sUAS) for performing a task while considering task complexity and environmental factors. Our framework is a cascaded Fuzzy Inference System (cFIS) composed of combination of three FIS which represent different contextual autonomy capabilities. We performed several experiments to test our framework in various contexts, such as endurance time, navigation, take off/land, and room clearing, with seven different sUAS. We introduce a predictive measure which improves upon previous predictive measures, allowing for previous real-world task performance to be used in predicting future mission performance.
\end{abstract}


\section{Introduction}
\label{intro}
In today's world, robots are expected to become increasingly present by assisting humans in performing various tasks in different environments. While some robots have been designed for a single purpose, others can accomplish a variety of tasks with different levels of autonomy. Measuring robot autonomy is an important and ever evolving concept and existing methods for evaluating robot autonomy can be categorized into two main families: contextual and non-contextual. While the former methods consider mission and task-specific measures (e.g., ALFUS~\cite{durst2014levels}, ACL~\cite{2002-Clough}), the latter only rely on implicit system capabilities and do not consider the mission and environment features (e.g., NCAP~\cite{durst2011non, hertel2022methods}).

Our study in this paper focuses on evaluating the contextual autonomy for small Unmanned Aerial Systems (sUAS). Existing methods such as ALFUS~\cite{durst2014levels} and MPP \cite{2014mpp} share a similar shortcoming in that neither provides a simple implementation for use with real-world systems. Another drawback of existing methods that our approach addresses is the lack of a consistent process for breaking down tasks into sub-tasks and combining scores calculated for sub-tasks into a unified score for the given task.
In this paper we propose a method for evaluating the contextual autonomy of sUAS based on a fuzzy interface that allows the operator to design and modify the evaluation system using linguistic reasoning. We designed four indoor tasks (endurance time, navigation, takeoff/land, and room clearing) and tested our interface in various experiments with seven different sUAS. Our results show that the proposed approach calculates a contextual autonomy score that can be used to rank the systems for each context.




\section{Related Work}
\label{related works}

Some of the first and more simplistic methods of categorizing autonomous systems are the Levels of Automation (LOA) proposed by Sheridan~\cite{1991-Sheridan} and its later expansion~\cite{2000-LOA}. LOA defines automation as ``the full or partial replacement of a function previously carried out by the human operator'' in a 1 to 10 range; 1 being full control by the human and 10 being full control by the computer. LOA does not accurately describe how outside factors can affect the autonomous capability of a system. While it could theoretically be applied to a robot, it would not be accurate as it fails to accommodate for differing degrees of difficulty in tasks, and environmental factors. 

Another evaluation method is known as the Autonomy Control Levels (ACL)~\cite{2002-Clough}. ACL is designed for Unmanned Aerial Vehicles (UAV), and operates on a similar basis of utilizing autonomy levels from 0-10, with 0 being fully remotely controlled by a pilot, and 10 being a human-like system. These levels closely resemble the 10 LOA, following the same concept.  The ACL characterizes each system according to four metrics, which attempt to categorize different areas of autonomous behaviors for the system. In each of these, an autonomy level from 0-10 is given based upon these behaviors. This system has a similar drawback, in that it does not account for difficulties in the mission itself.


Another method is the Autonomy and Technological Readiness Assessment (ATRA)~\cite{kendoul2013towards}. ATRA attempts to combine both the basic theory behind the Autonomy Level, and the Technology Readiness Level (TRL) metric into one framework~\cite{kendoul2013towards}. 
TRL utilizes these two metrics in an attempt to evaluate the autonomy level provided by different technologies onboard the UAS. This is emphasized as a solution for the gap between existing theoretical work and technological advances in the UAS autonomy space.

Autonomy Levels for Unmanned Systems (ALFUS) is a method for defining the autonomy of a system in terms of three different axes~\cite{huang2007autonomy}. ALFUS has a strong theoretical basis, but somewhat impractical in the real-world implementation due to the lack of maturity in some of these systems, as well as the inability of most, if not all, available systems to reach the upper levels of the three axes.

The three axes mentioned are known as the Mission Complexity (MC), Environmental Complexity (EC), and Human Independence (HI) axes. Each one of these axes pertains itself to a different aspect of the contextual autonomy of a system. The MC axis pertains mostly towards the difficulty of the tasks and movements required of the system to complete the task (e.g. maneuvers, speed, searching). Alternatively, the EC axis concerns itself with the difficulty in the performance of the task caused by environmental factors (e.g. Lighting, Obstacles, Enclosed Spaces). Lastly, the HI axis is representative of the level of independence between the user and the system (e.g. task planning, task execution). 

Due to the ability to split the representation of a system's autonomy into these three axes, it allows for the characterization and evaluation of system's autonomy in real world tests, including the impact that both the environment and the mission profile can have on the system's autonomy. Our work in this paper is based off many of the ideas put forward through ALFUS, and we utilize it as a foundational part of our contextual autonomy evaluation.

The Mission Performance Potential is proposed as a method for the evaluation of a unmanned system's autonomous performance, as well as a predictor for future missions~\cite{2014mpp}. This method provides a metric which represents the maximum performance of a system in a given mission at a given autonomy level. Uniquely, this method includes both non-contextual autonomy metrics, and contextual autonomy metrics, and provides a single output prediction based on both types of data.

One of the drawbacks of MPP is that it only provides a prediction of the performance of a system at a specified autonomy level for a specified mission. In other words, this does not evaluate how a real system performs, but rather the maximum potential for a system to perform. Our approach instead calculated the actual autonomy of a system based on actual data from real-world experiments.

\begin{figure}[htbp]
    \centering
    \includegraphics[trim=0 1.5cm 0 0.5cm, clip, width=0.9\linewidth]{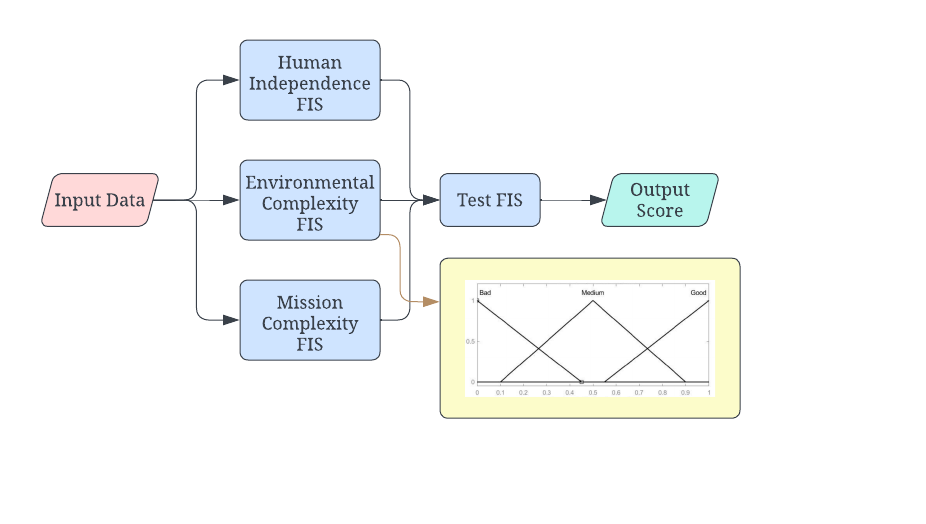}
    \caption{Our cascaded Fuzzy Inference System used for calculating a contextual autonomy score for a performed task.}
    \label{fig:my-system}
\end{figure}

\section{Framework}
\label{framework}

ALFUS' summary model works with a set of metrics for each of its three axes, as well as a system of levels from 0 to 10. These levels are based upon possible answers from those metrics, to provide a level evaluation of a system. As a generic framework, ALFUS tends to have a very broad, and somewhat open to interpretation, definition of metrics. For instance, in the case of the EC axis,  it ranges from a ``simple environment,'' to an ``extreme environment.'' However, the summary model describes the system in terms of an autonomy level for each axis, while the Contextual Autonomy Capability within ALFUS provides an actual score for each axis. Due to the autonomy level evaluation, there is some ambiguity when characterizing systems. This is one of the main concerns with ALFUS, in that while it does provide a strong theoretical background, the actual implementation of the ideas with real-world systems is not as clear.

We utilize Takagi-Sugeno Fuzzy Inference Systems (FIS) as a means to combine different metrics in an evaluation of an sUAS which is both easy to use, and allows us to use some data which is either not easily defined numerically, or inherently qualitative about the environment, combined with standard quantitative metrics. Fuzzy inferences also allow for slight deviations in a metric to not cause a drastic change in the evaluation of that sUAS. We designed a set of tests with various mission and environment complexity levels (see Section~\ref{tests}), and defined a fuzzy inference system for each test. Unlike MPP~\cite{2014mpp}, our fuzzy inference systems are based on the three-axis model used in ALFUS, by creating an individual FIS for metrics associated with each axis (i.e., MC, EC, HI), and an additional FIS which combines these three outputs into a single score. This structure representing a cascaded FIS (cFIS) is illustrated in Fig.~\ref{fig:my-system}. For each test, the outcome of the FIS for all three axis is fed into a combining FIS that produces a final autonomy score. Each FIS in our cFIS is a Sugeno-type FIS with multiple inputs and one output. For each input of an FIS, we consider three membership functions (MFs) labeled as \emph{low}, \emph{medium}, and \emph{high}. Without loss of generality, we used  triangular MFs, however, other types of MF can be used. The input variables used in different tests and their corresponding MF parameters have been reported in Table~\ref{tab:var-mf}.  The output of each Sugeno-type FIS has five singleton MFs (i.e., constant):  \emph{very bad}, \emph{bad}, \emph{medium}, \emph{good}, \emph{very good}. Our FIS' use a triangular fuzzifier and a Sugeno defuzzifier (i.e., weighted average output). For each FIS, we defined a rule base (i.e., a set of linguistic rules).

In the cFIS structure in Fig.~\ref{fig:my-system}, the defuzzified output of each FIS is a value in the range of $[0,1]$. For the initial three FIS, $0$ and $1$ represent the lowest and highest complexity, respectively. In the case of the final FIS, $0$ and $1$ represent the lowest and the highest autonomy, respectively. If we define the singleton value of each output function as $z_i$, and the degree to which each output is weighted based upon the ruleset as $w_i$, then the output final score can be calculated as follows:
\begin{equation}
    s = \frac{\sum_{i=1}^{N}w_{i}z_{i}}{\sum_{i=1}^{N}w_{i}}
\end{equation}
\noindent where $N$ represents the number of rules in the rule base. Table~\ref{tab:fuzzy-rules} reports an example of the fuzzy ruleset we used. The advantage of this system is that we can utilize many different types of data, and clearly define the ranges for each value, allowing the pilots performing the tests to provide feedback on the membership functions and rulesets.

\begin{table}[htbp]
    \centering
    \begin{tabularx}{0.88\linewidth}{ccccc} \cline{3-5}
    & & \multicolumn{3}{c}{Mission Complexity Axis} \\ 
    & & Low & Medium & High \\ \cline{1-5}
    \multirow{3}{*}{\shortstack{Environment \\ Complexity \\ Axis}} & \multicolumn{1}{l|}{Low} & Very Bad & Bad & Medium  \\
    & \multicolumn{1}{l|}{Medium} & Bad & Medium & Good \\
    & \multicolumn{1}{l|}{High} & Medium & Good & Very good \\\bottomrule \hline

    \end{tabularx}
    \caption{\small{Fuzzy Ruleset utilized in our final combinational FIS}}
    \label{tab:fuzzy-rules}
\end{table}

\section{UAS Platforms}
\label{platforms}

\begin{figure}[t]
\centering
\minipage{.14\textwidth}
	\includegraphics[width=\linewidth]{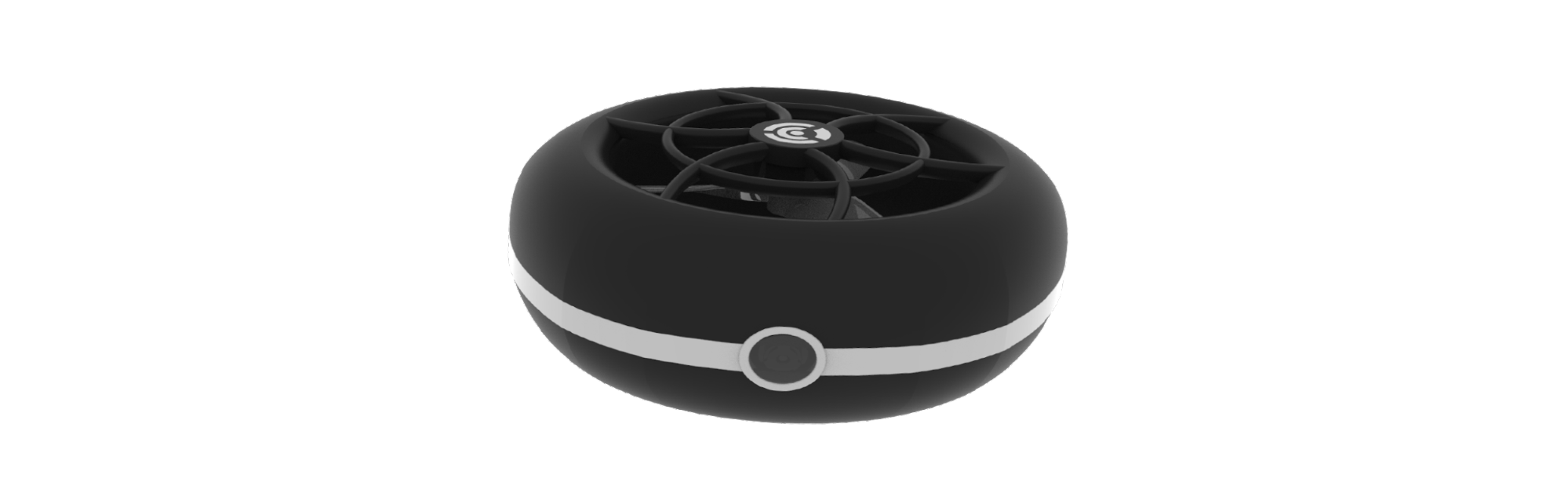}
\endminipage\hfill
\minipage{.14\textwidth}
	\includegraphics[width=\linewidth]{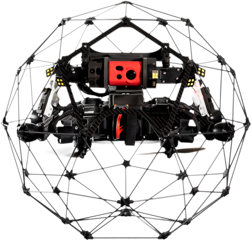}
\endminipage\hfill
\minipage{.14\textwidth}
	\includegraphics[width=\linewidth]{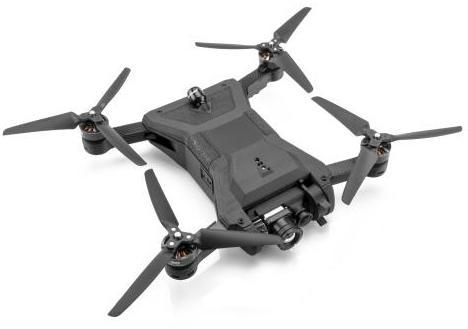}
\endminipage\hfill
\minipage{.14\textwidth}
	\includegraphics[width=\linewidth]{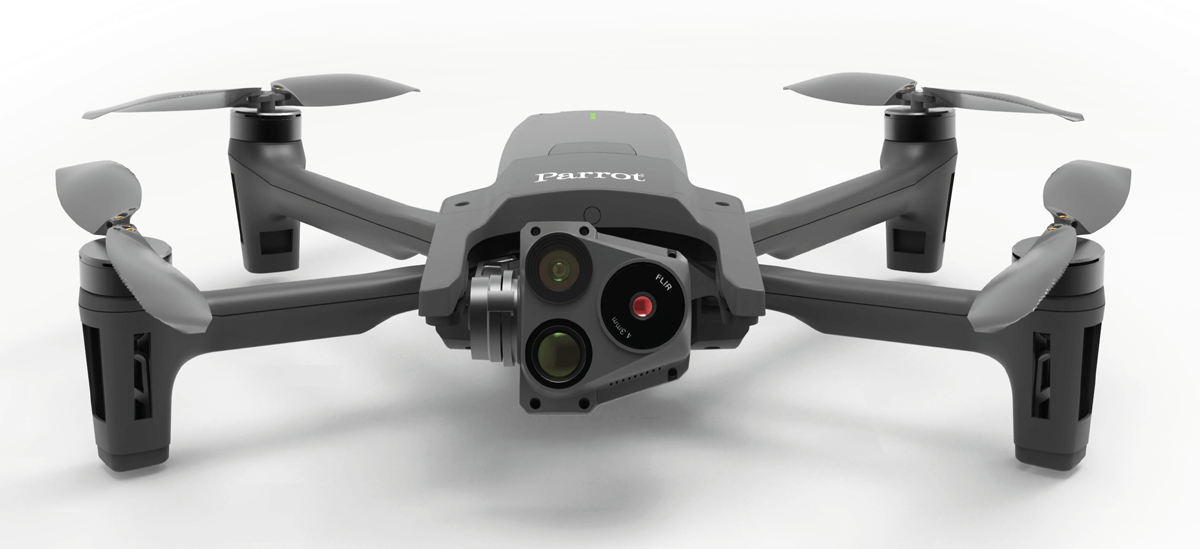}
\endminipage\hfill
\minipage{.14\textwidth}
	\includegraphics[width=\linewidth]{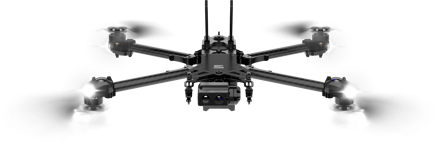}
\endminipage\hfill
\minipage{.14\textwidth}
	\includegraphics[width=\linewidth]{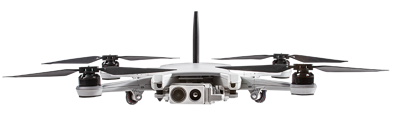}
\endminipage\hfill
\minipage{.14\textwidth}
	\includegraphics[width=\linewidth]{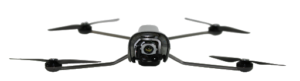}
\endminipage\hfill
\caption{\small{From left to right, top to bottom: Cleo Robotics Dronut, Flyability Elios 2, Lumenier Nighthawk 3, Parrot ANAFI USA GOV, Skydio X2D, Teal Drones Golden Eagle, Vantage Robotics Vesper}}
\label{fig:platforms}
\end{figure}

Fig.~\ref{fig:platforms} illustrates seven sUAS platforms evaluated in our experiments. The platforms include: the Cleo Robotics Dronut\footnote{\url{https://cleorobotics.com/}}, Flyability Elios 2\footnote{\url{https://www.flyability.com/elios-2}}, Lumenier Nighthawk 3\footnote{\url{https://www.lumenier.com/}}, Parrot ANAFI USA GOV\footnote{\url{https://www.parrot.com/us/drones/anafi}}, Skydio X2D\footnote{\url{https://www.skydio.com/skydio-x2}}, Teal Drones Golden Eagle\footnote{\url{https://tealdrones.com/suas-golden-eagle/}}, and Vantage Robotics Vesper\footnote{\url{https://vantagerobotics.com/vesper/}}. These platforms provide a wide ranging set of capabilities and use cases. For instance, Parrot, Skydio X2D, Golden Eagle, and Vesper were developed for outdoor reconnaissance, whereas the Dronut and Elios 2 were developed for indoor reconnaissance and inspection, specifically in urban and industrial environments. Previously, we have used the same set of sUAS for a non-contextual benchmarking~\cite{hertel2022methods, norton2022decisive}. In our evaluations, we have anonymized the data by assigning the platforms labels A through G without any specific ordering or correlation. 


\begin{figure*}[ht]
\begin{minipage}{0.48\textwidth}
\begin{subfigure}{\linewidth}
\includegraphics[width=\linewidth,height=0.9in]{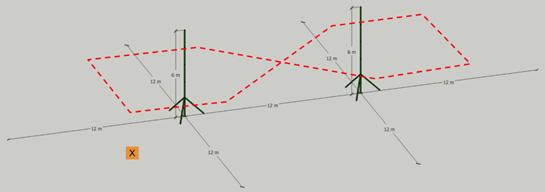}
\caption{Runtime Endurance test showing a system performing a specified movement} \label{subfig:test-runtime}
\end{subfigure}
\begin{subfigure}{\linewidth}
\includegraphics[width=\linewidth, height=2in]{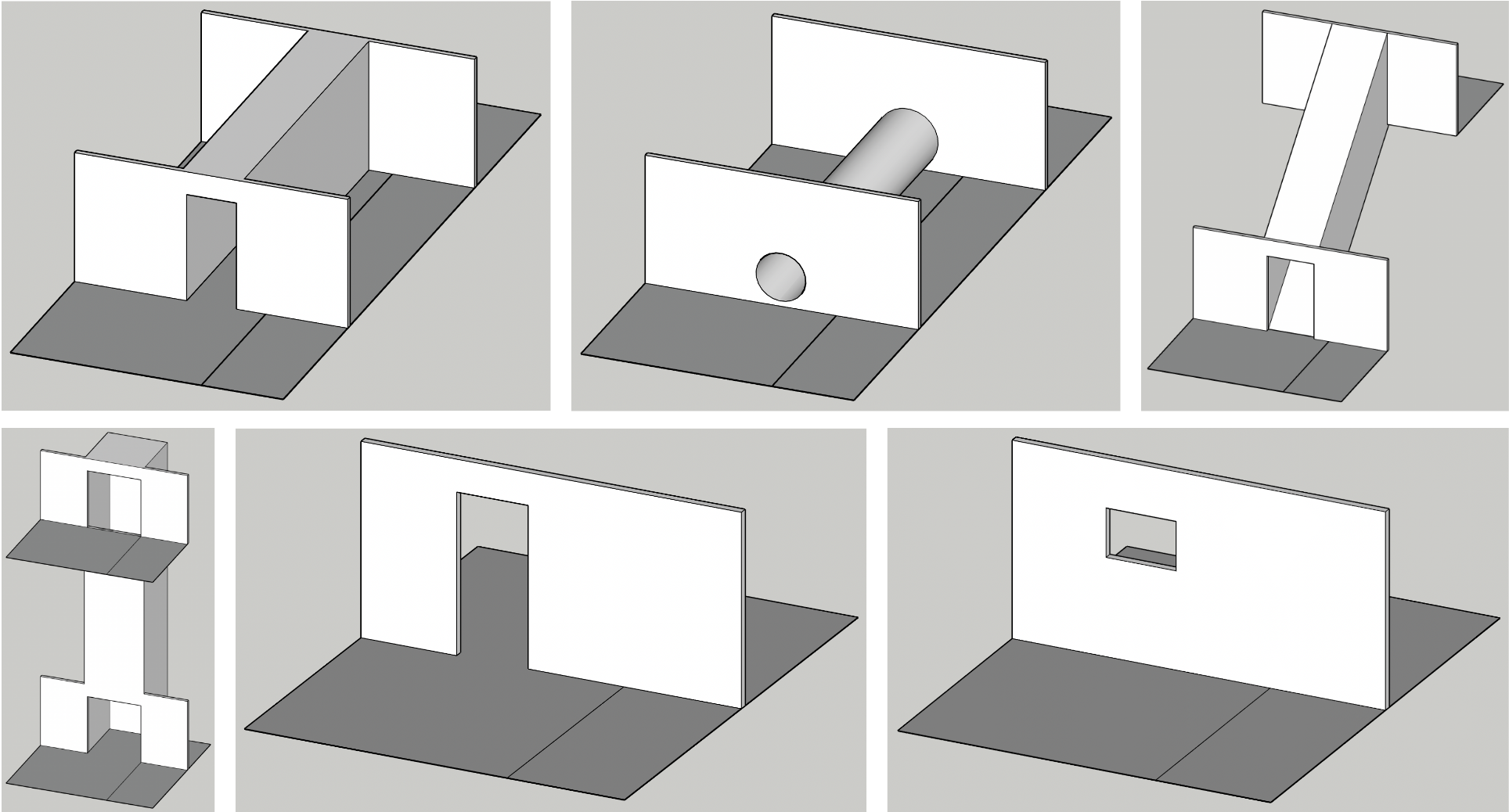}
\caption{Navigation tests (left to right, top to bottom): hallway, tunnel, stairwell, shaft, door, window.} \label{subfig:test-navigation}
\end{subfigure}
\end{minipage}
\begin{minipage}{0.48\textwidth}
\begin{subfigure}{\linewidth}
\includegraphics[width=\linewidth,height=0.9in]{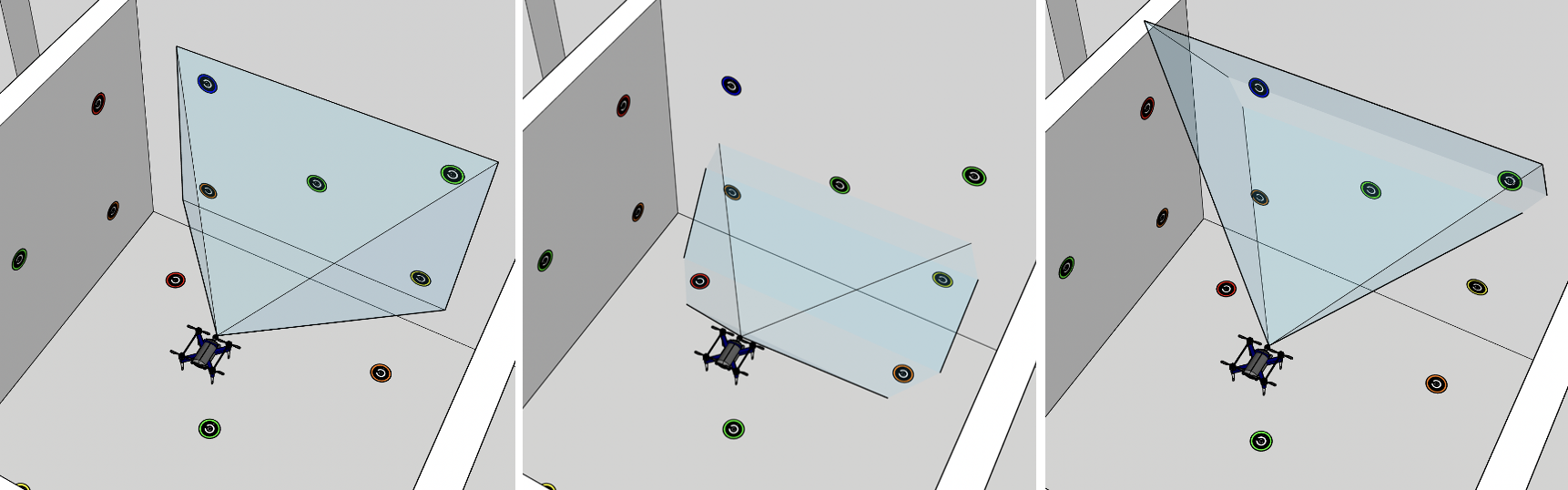}
\caption{Room clearing test showing a system inspecting surfaces for visual acuity targets.} \label{subfig:test-roomclearing}
\end{subfigure}
\begin{subfigure}{\linewidth}
\includegraphics[width=\linewidth,height=2in]{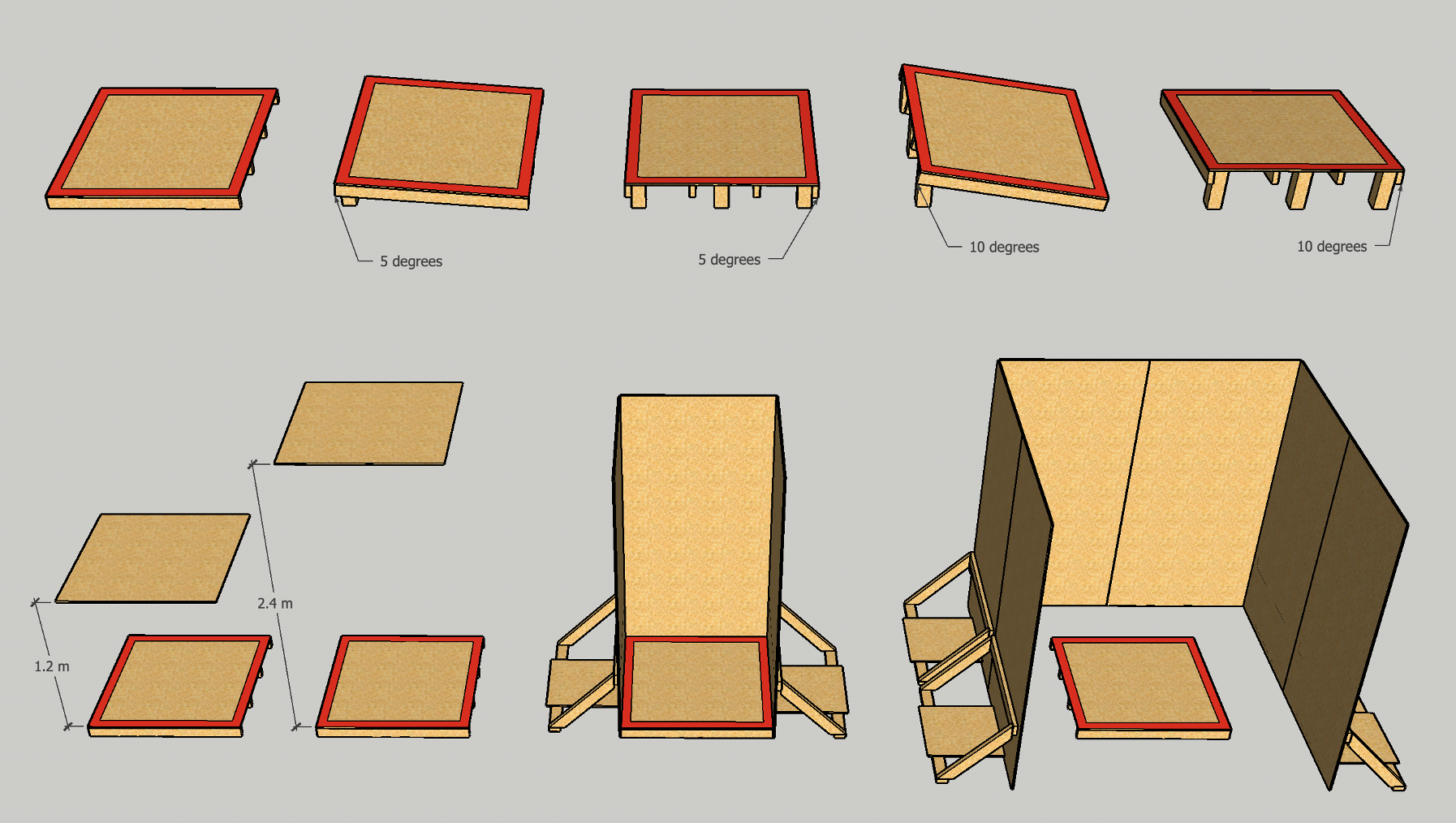}
\caption{Takeoff and land/perch tests showing variations in ground plane angle (top row) and nearby obstructions (bottom row).} \label{subfig:test-takeoff}
\end{subfigure}
\end{minipage}
\caption{Tests designed for the evaluation of sUAS contextual autonomy.}
\label{fig:tests}
\end{figure*}
\section{Test Design}
\label{tests}

To evaluate the contextual autonomy of our platforms, we have designed several tests across a spectrum of areas. The variables for which we collected data for each test is reported in Table~\ref{tab:var-mf}. In this section, we describe each test briefly. As shown in Fig.~\ref{fig:tests} all tests have been designed for indoor environments.

\subsection{Runtime Endurance}
This family of tests focuses on the battery life of the system in various operational profiles. As shown in Fig.~\ref{subfig:test-runtime}, the specific test we use from this group focuses on the system flying continuously in a figure-8 
pattern. The main performance metric for the test is the test duration.

\subsection{Navigation}
We have designed two main types of navigation tests, each with several profiles defined based on the type of movement (horizontal, vertical, or both) and the type of confinement (horizontal, vertical, or both). As shown in Fig.~\ref{subfig:test-navigation}, navigation through confined spaces involves traversal into and out of a continuously confined space, with tests for hallway, tunnel, stairwell, and shaft. Navigation through apertures involves transient traversal through an opening, with tests for doorway and window. Each navigation environment is characterized according to the dimensions of the confined space or aperture, lighting, surface textures, and the presence of obstructions on either side of the confined space or aperture. The main metrics of performance are efficacy and average navigation time.


\subsection{Room Clearing}
In this test method, the system performs a visual inspection of an example room whose walls, floor, and ceiling are outfitted with visual acuity targets which contain nested Landolt C symbols of decreasing size. As shown in Fig.~\ref{subfig:test-roomclearing}, the test was performed under two conditions: with and without using camera zoom. The main performance metrics are duration, coverage, and average acuity.

\subsection{Takeoff and Land/Perch}
As illustrated in Fig.~\ref{subfig:test-takeoff}, these tests evaluate the system's ability to takeoff and land or perch in various environments that may be affected by stabilization issues or preventative safety checks from the system. The conditions tested vary the angle of the ground plane (flat, $5\degree$ and $10\degree$ pitch and roll) and the presence of obstructions ($1.2$-$2.4$m overhead, $0.6$-$1.2$m lateral). The main performance metric is efficacy.

\section{Results}
Utilizing our framework outlined in Section~\ref{framework}, we calculate a performance score for each sUAS based upon the conditions and performance metrics detailed below. As mentioned above, our system provides a single score for each of the three attributes, the EC, MC and HI axes, and utilizes those scores to provide a single score for the entire test. It should be noted that although we consider all three axes in our cFIS structure, due to lack of data, we consider the lowest level (i.e., full tele-operation) for the HI axis across all tests. The test-specific details of the structure is discussed in corresponding sections. Another factor to note is that for some of these experiments, some sUAS were not available at the time of testing, due to the sUAS being repaired, or other circumstances. In this case, we attempt to remedy this by calculating a partial point achieved by the sUAS. For situations where data for an entire test is missing, we cannot fully evaluate the sUAS. However, we have evaluated sUAS for individual tests for which the data was recorded. Despite these edge cases, most sUAS had available data, which was used in our evaluations.

\begin{figure}[htbp]
\begin{minipage}{0.48\linewidth}
\begin{subfigure}{\linewidth}
\includegraphics[width=\linewidth, height=1.25in]{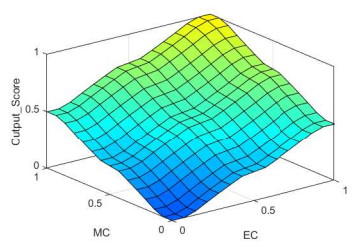}
\caption{Final FIS Surface} \label{subfig:final-surface}
\end{subfigure}

\begin{subfigure}{\linewidth}
\includegraphics[width=\linewidth, height=1.25in]{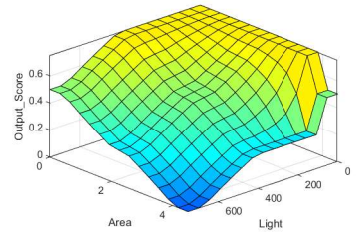}
\caption{Through Aperture test EC} \label{subfig:aperture-surface}
\end{subfigure}

\end{minipage}
\begin{minipage}{0.48\linewidth}
\begin{subfigure}{\linewidth}
\includegraphics[width=\linewidth, height=1.25in]{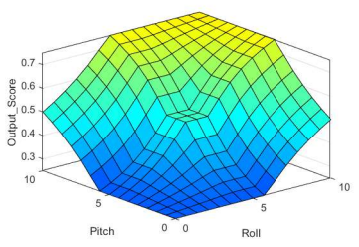}
\caption{Takeoff and Land/Perch EC} \label{subfig:takeoff-surface}
\end{subfigure}
\begin{subfigure}{\linewidth}
\includegraphics[width=\linewidth, height=1.25in]{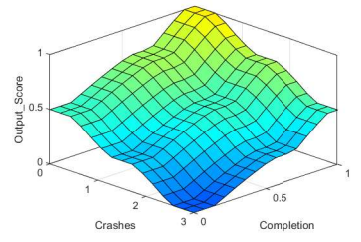}
\caption{Through Corridor MC} \label{subfig:corridor-surface}
\end{subfigure}
\end{minipage}

\caption{FIS surfaces for different tests.}
\label{fig:surfaces}
\end{figure}
\begin{figure}[htbp]
\centering
\includegraphics[width=\linewidth]{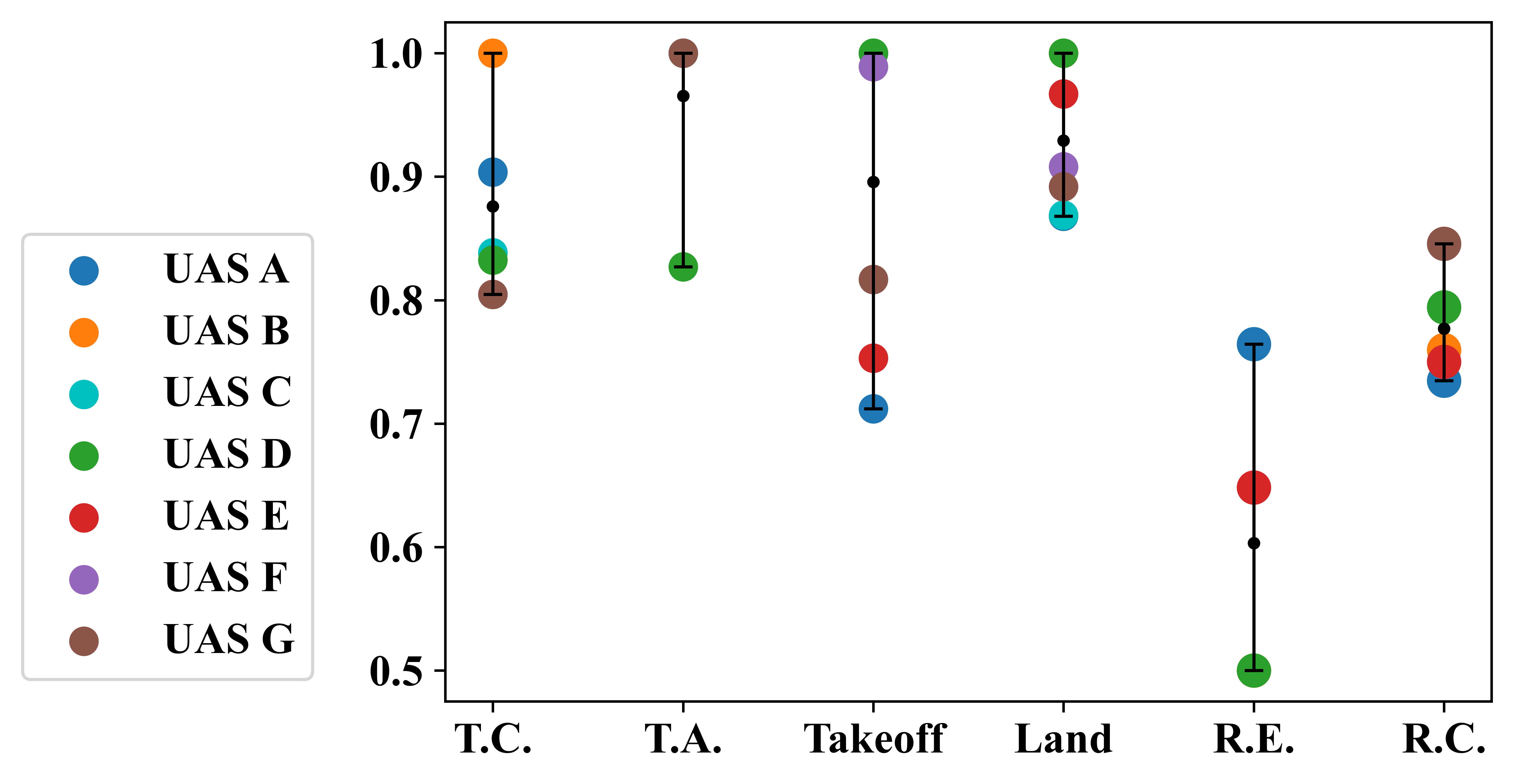}
\caption{\small{Scores for each sUAS as a percentage of the maximum score possible on the y-axis, with each test on the x-axis}}
\label{fig:score_tables}
\end{figure}

As mentioned before, the proposed structure in Fig.~\ref{fig:my-system}, was adapted to each specific test. The resulting cascaded FIS are depicted in Fig.~\ref{fig:cascaded-fis} each of which represents a cFIS for a specific test. More specifically, each sub-figure shows the inputs and outputs of each cFIS, as well as how the FIS modules are connected. This is meant to provide a visual aid, which can be useful to keep track of each FIS, as we discuss the results. Associated membership functions for each input, and the outputs, can be found in Table~\ref{tab:var-mf}. Some of the FIS surfaces, which show the relationship between two input values in an FIS, and the corresponding output value, are shown in Fig.~\ref{fig:surfaces}. Additionally, numerical results are reported in Fig.\ref{fig:score_tables} and Table~\ref{tab:final-results}. 

\begin{table*}
\scriptsize
\centering
\begin{tabularx}{\linewidth}{@{} *5l @{}}    \toprule
\emph{Variables} & \emph{Description} & &\emph{MFs: Triangular\{Low, Medium, High\} } & \\\midrule
Area    & Aperture/Hallway Cross-Section ($m^2$)  & $[0, 0, 2.7]$ &$[0.6, 3, 5.4]$& $[3.3, 6, 6]$ \\ 
Light & Ambient Light Level (Lux) & $[0,0,337.5]$ & $[75,375,675]$ & $[412.5,750,750]$\\ 
Vert & Verticality ($\degree$) & $[0,0,37.5]$ & $[7.5,45,82.5]$ &$[52.5,90,90]$\\
Crash & Number of Crashes & $[0,0,1.25]$ & $[0.5,1.5,2.5]$ & $[1.75,3,3]$\\
Rollovers & Number of Rollovers & $[0,0,1.25]$ & $[0.5,1.5,2.5]$ & $[1.75,3,3]$\\
Comp. \% & Completion Percentage & $[0,0,0.55]$ & $[0.15,0.6,0.92]$ & $[0.7,1,1]$\\
Yaw/Pitch & Static Yaw/Pitch Angle ($\degree$) & $[0,0,4.17]$ & $[0.83,5,9.12]$ & $[5.83,10,10]$\\
VR & Static Vertical Obstruction ($m$) & $[0.6,0.6,1.1]$ & $[0.7,1.2,1.7]$ & $[1.3,1.8,1.8]$\\
LR & Static Lateral Obstruction ($m$) & $[1.2,1.2,2.2]$ & $[1.4,2.4,3.4]$ & $[2.6,3.6,3.6]$\\
Coverage & Coverage Percentage & $[0,0,0.55]$ & $[0.15,0.6,0.92]$ & $[0.7,1,1]$\\
Cs Detected & Landolt C Depth Detected & $[0, 0, 50]$ & $[10, 50, 90]$ & $[50, 100, 100]$\\
Duration & Duration of Test (Minutes) & $[2.5, 2.5, 5.25]$ & $[3.05, 5.25, 7.45]$ & $[5.25, 8, 8]$\\
Obs. & Number of Obstructions & $[0,0,2.5]$ & $[1,3,5]$ & $[3.5,6,6]$\\

\midrule
\emph{Output Variable} & \emph{Description} & &\emph{Sugeno MFs: Constant \{Very Low to Very High\} } & \\
\midrule
Score & Combined Defuzzified Score &  & $[0, 0.25, 0.5, 0.75, 1]$ &\\ 
\bottomrule
 \hline
\end{tabularx}
\caption{\small{Membership Functions (MFs) for each input and output variable used in an FIS in the evaluation of these sUAS.}}
\label{tab:var-mf}
\end{table*}

\subsection{Test Results}

\begin{figure}[htbp]
\begin{minipage}{0.48\textwidth}
\begin{subfigure}{\linewidth}
\includegraphics[width=\linewidth, height=1.in]{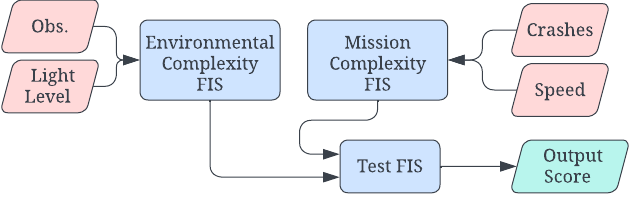}
\caption{\small{Runtime Endurance (R.E.) test FIS}} \label{subfig:fis-runtime-endurance}
\end{subfigure}
\begin{subfigure}{\linewidth}
\includegraphics[width=\linewidth, height=1.in]{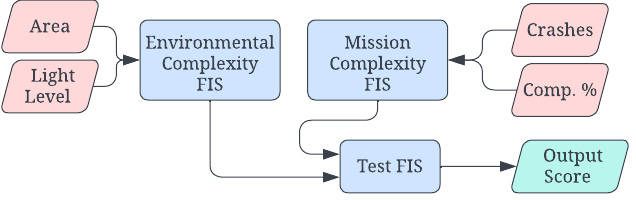}
\caption{\small{Through Apertures (T.A.) test FIS}} \label{subfig:fis-through-apertures}
\end{subfigure}
\begin{subfigure}{\linewidth}
\includegraphics[width=\linewidth, height=1.2in]{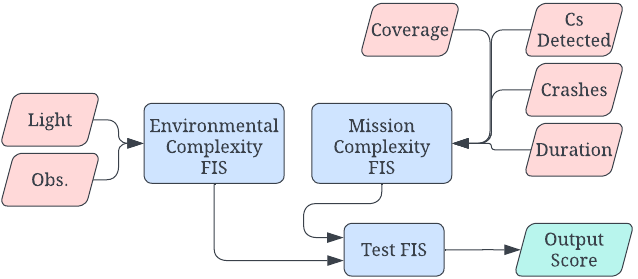}
\caption{\small{Room Clearing (R.C.) test FIS}} \label{subfig:fis-room-clearing}
\end{subfigure}
\begin{subfigure}{\linewidth}
\includegraphics[width=\linewidth, height=1.in]{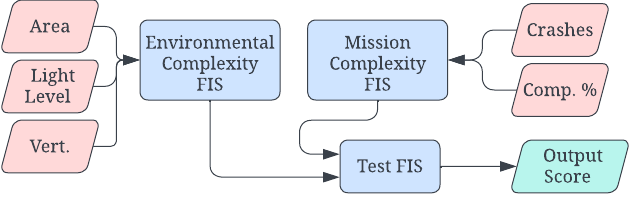}
\caption{\small{Through Corridors (T.C.) test FIS}} \label{subfig:fis-through-corridors}
\end{subfigure}
\begin{subfigure}{\linewidth}
\includegraphics[width=\linewidth, height=1.2in]{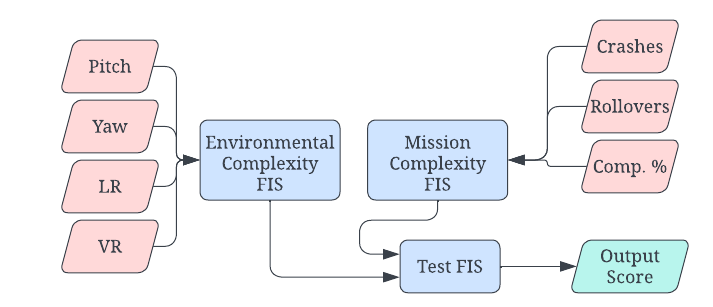}
\caption{\small{Takeoff and Land/Perch test FIS}} \label{subfig:fis-takeoff-land}
\end{subfigure}
\end{minipage}

\caption{Diagrams of each system of cascaded FIS utilized to calculate scores for each test}
\label{fig:cascaded-fis}
\end{figure}

\textbf{Runtime Endurance:} The runtime endurance test is likely the simplest of the tests performed, however, it is still a useful test in gauging how a sUAS might perform in a real mission. Fig.~\ref{subfig:test-runtime} illustrates the runtime endurance test design including the navigation path and two stands. The adapted cFIS for this test is shown in Fig.~\ref{subfig:fis-runtime-endurance} with four inputs: number of obstructions, number of crashes, light level and speed.

Despite the simplicity of this test, some sUAS did not performed well  largely due to slow speed. Our results indicate that some sUAS may have trouble in portions of this real-world mission which requires both speed and maneuverability. In should be noted that during this test only four sUAS were available.

\textbf{Navigation:} 
In the navigation tests, due to the differences between the \emph{through corridor} and \emph{through apertures} tests, two slightly different cascaded FIS were designed. These can be found in Fig.~\ref{subfig:fis-through-apertures} and Fig.~\ref{subfig:fis-through-corridors}. The input to the \emph{through corridors} cFIS includes area (cross section), light level, verticality, coverage, number of crashes, and duration. The inputs to the \emph{through apertures} cFIS include area, light level, number of crashes, and completion percentage.
As shown above in Fig.~\ref{fig:score_tables}, each sUAS that performed the \emph{through apertures} test, achieved a maximum score, besides sUAS G, which performed slightly worse, due to both issues in correctly traversing the aperture, as well as being the only sUAS to suffer a crash during the test. 
Next, for the \emph{through corridors} test, As shown in Fig.~\ref{fig:score_tables}, there is more variance in the performance between each sUAS, with UAS A performing the best, and UAS E performing the worst, even though sUAS E tied for best in \emph{through apertures}. This is important, as there is more room for error while traversing corridors, than there is traversing an aperture. 

\textbf{Takeoff and Land/Perch:} 
For the takeoff and land/perch test, the cFIS diagram can be found in Fig.~\ref{subfig:fis-takeoff-land}. The inputs include pitch, yaw, number of crashes, completion percentage, vertical and lateral obstruction, and number of rollovers. As can be seen in Fig.~\ref{fig:score_tables} and Table~\ref{tab:final-results}, sUAS A and sUAS G perform best across both sections of the test and thus provide the highest level of autonomy. Likewise, sUAS B performs the worst in both portions of the test, showcasing a lower level of autonomy, compared to the other sUAS. This evaluation allows for the characterization of how an sUAS may perform during portions of a mission which requires the system to takeoff or land in a specified spot, of varying difficulty.

\textbf{Room Clearing:} Since the room clearing test is done in a static environment, we included a time constraint in our testing making the evaluation to focus on the performance of the sUAS in regards to the Mission Complexity axis.
The designed cFIS can be seen in Fig.~\ref{subfig:fis-room-clearing}. Inputs include light level, number of obstructions, number of crashes, duration, coverage, and landolt C depth detected. Results are found in Fig.~\ref{fig:score_tables}, and Table~\ref{tab:final-results}.
In this test, the strongest performer was UAS E; however, most of the UAS performed closely to each other. Surprisingly, a strong performance in the runtime endurance test did not necessarily correlate to a strong performance in this test. Both of these tests require a system with good maneuverability capabilities, but this test also requires a controller which allows the user to visually identify different landmarks.

\subsection{Final Results}

\begin{table}[htbp]
    \centering
    \begin{tabularx}{\linewidth}{cccccccc} \toprule
    \emph{UAS} & \emph{T.C.} & \emph{T.A.} & \emph{Takeoff} & \emph{Land} & \emph{R.E.}  & \emph{R.C.} & \emph{\shortstack{Predictive \\ Score}} \\\midrule
    A & 1.0 & 1.0 & 1.0 & 1.0 & 0.5 & 0.76 & 0.85 \\
    B & 0.90 & 1.0 & 0.71 & 0.87 & 0.76  & 0.73 & 0.82 \\
    C & 0.84 & 1.0 & 1.0 & 0.87 & -  & - & 0.92 \\
    D & - & - & 0.75 & 0.97 & 0.65  & 0.75 & 0.77 \\
    E & 0.80 & 1.0 & 0.82 & 0.89 & -  & 0.85 & 0.87 \\
    F & - & - & 0.99 & 0.91 & -  & - & 0.95 \\
    G & 0.83 & 0.83 & 1.0 & 1.0 & 0.5  & 0.79 & 0.80 \\
    \bottomrule
    \hline
    
    \end{tabularx}
\caption{\small{Scores of each sUAS for each test, as well as a weighted multiple which allows for an overall  evaluation of each sUAS}}
\label{tab:final-results}
\end{table}

Contextual autonomy evaluations are concerned with the performance of a system within an environment while performing a specific task with a known level of complexity. However, calculating an overall score that represents an average autonomy for a given system in a spectrum of tests and environment is desirable. To combine the test scores into a single score, we utilize a weighted product, with equal weightings for each test, as we did previously in our non-contextual evaluation\cite{hertel2022methods}. The weighted product represented as 
\begin{align}
    P = \prod_i^M \phi_i^{w_i},
\end{align}
\noindent where $M$ is the number of individual tests, $\phi_i$ represents an individual test score, and $w_i$ the weight assigned to that test. has several benefits including that different test results can be combined without requiring normalization or scaling. The results for each sUAS are shown in Table~\ref{tab:final-results}. It is important to note that many of the sUAS perform better than the others in some tests, but worse in others. One example of this is UAS A, which has the fourth highest weighted multiple (overall score), while performing the best in four out of six tests. This is due to the sUAS's relatively poor performance in both the runtime endurance test, as well as in the room clearing test. 

The use case of a singular score like this presents itself when a user would like to know which sUAS is likely to provide the most overall autonomy, across multiple tests and different environment.

\section{Conclusions and Discussions}

In this paper, we proposed a framework for evaluation of contextual autonomy for robotic systems. Our framework consists of a cascaded Fuzzy Inference System (cFIS) that combines test results over three axes of evaluation (mission complexity, environment complexity and human independence) introduced by the ALFUS framework. We have designed four tests with different mission complexity and environment complexity levels and performed several experiments with several sUAS, and we have shown that our modular framework is adaptable to different tests. For future work, we plan to extend our framework for performance evaluation.

To achieve this, a desired mission can be decomposed into base tasks, such as takeoff/landing, traversing through environments/apertures, clearing rooms, and general maneuverability. The user then can define a set of weights and calculate a potential performance score of a sUAS for the target mission. Unlike MPP~\cite{2014mpp}, however, we suggest a method which is based upon performance in set tasks, rather than a combination of non-contextual attributes, and environmental factors. 


\section*{Acknowledgment}
This work is sponsored by the Department of the Army, U.S. Army Combat Capabilities Development Command Soldier Center, award number W911QY-18-2-0006. Approved for public release \#PR2022\_88282

\bibliographystyle{IEEEtran}
\bibliography{references}

\end{document}